\documentclass{acm_proc_article-sp}
\usepackage{amsmath}
\usepackage[utf8x]{inputenc}                    
\usepackage[portuguese,english]{babel}
\usepackage{dblfloatfix}

\newcommand{\E}{\mathop{\mathbb E}}

\begin{document}

\title{A Labeled Graph Kernel for Relationship Extraction}

\numberofauthors{3} 
\author{
\alignauthor
Gonçalo Simões\\
\affaddr{INESC-ID / IST, PT}\\
\email{goncalo.simoes@ist.utl.pt} 
\alignauthor
David Matos\\
\affaddr{INESC-ID / IST, PT}\\
\email{david.matos@inesc-id.pt} 
\alignauthor
Helena Galhardas\\
\affaddr{INESC-ID / IST, PT}\\
\email{helena.galhardas@ist.utl.pt}
}

\maketitle

{
\begin{abstract}
In this paper, we propose an approach for Relationship Extraction (RE) based on labeled graph kernels. The kernel
we propose is a particularization of a random walk kernel that exploits two properties previously studied in the
RE literature: \textit{(i)} the words between the candidate entities or connecting them in a syntactic representation
are particularly likely to carry information regarding the relationship; and \textit{(ii)} combining information from
distinct sources in a  kernel may help the RE system make better decisions. We performed experiments on a dataset of
protein-protein interactions and the results show that our approach obtains effectiveness values that are comparable with the
state-of-the art kernel methods. Moreover, our approach is able to outperform the state-of-the-art kernels when combined
with other kernel methods.
\end{abstract}




\section{Introduction}

With the increasing use of Information Technologies, the amount of unstructured text available in digital data
sources (e.g., email communications, blogs, reports) has grown at an impressive rate. These texts
may contain vital knowledge to Human decision making processes. However, it is unfeasible for a human to
analyze big amounts of unstructured information in a short time. In order to solve this problem, a typical approach
is to transform unstructured information in digital sources into a previously defined structured format.

Information Extraction (IE) is the scientific area that studies techniques to extract semantically relevant
segments from unstructured text and represent them in a structured format that can be understood/used by
humans or programs (e.g., decision support systems, interfaces for digital libraries). In the past few years,
there has been an increasing interest in IE, from industry and scientific communities. In fact, this
interest led to huge advances in this area and several solutions were proposed in applications such as Semantic
Web \cite{IESW} and Bioinformatics \cite{Giuliano06exploitingshallow,citeulike:3115793}.

Regardless of the application domain, an IE activity can be modeled as a composition of the following high-level tasks
\cite{McCallum:2005}:

\begin{itemize}
\item \textbf{Segmentation:} divides the text into atomic segments (e.g., words).
\item \textbf{Entity recognition:} assigns a class (e.g., organization, person) to each segment of the text.
Each pair (\textsf{segment}, \textsf{class}) is called an entity.
\item \textbf{Relationship extraction:} determines relationships (e.g., \textit{born\_in}, \textit{works\_for})
between entities.
\item \textbf{Entity normalization:} converts entities into a standard format (e.g., convert all dates to a
pre-defined format).
\item \textbf{Co-reference resolution:} determines which entities represent the same object/individual in the real world
(e.g., IBM is the same as ``Big Blue'').
\end{itemize}

In the last decade, several techniques to increase the accuracy of these tasks
were proposed. In this paper, we focus only on the Relationship Extraction (RE) task.
The approaches that are typically used for RE can be divided into two major groups: \textit{(i)} handcrafted solutions, in which 
the programs are manually specified by the user through a set of rules; and \textit{(ii)} Machine Learning solutions, in 
which the programs are automatically generated by a machine either by explicitly producing rules or by generating 
a statistical model that is able to produce extraction results with regard to a set of characteristics of the input 
text.

Most of the first approaches for RE were based on handcrafted rules \cite{Aone98sra:description,Humphreys98universityof}. 
Typically, they exploited common patterns and heuristics to extract the desired relationships from the results of complex 
Natural Language Processing chains. These solutions were able to produce good results in several specific domains. 
However, they need a lot of human effort to produce rules for distinct domains.

To overcome this problem of handcrafted solutions, the application of Machine Learning to
RE started to receive a lot of attention. Typically, machine learning techniques used for RE are supervised. However,
some works have exploited semi-supervised \cite{Brin98extractingpatterns,375774,Etzioni04web-scaleinformation,prdualrank} and 
unsupervised \cite{Eichler08g.:unsupervised, reVerb} techniques. Supervised approaches to RE are typically based on
classifiers that are responsible for determining whether there is a relationship or not between a set of entities.

There are two major lines of works in supervised approaches to RE: \textit{(i)} feature-based methods, which try
to find a good set of features to use in the classification process; and \textit{(ii)} kernel methods, which try
to avoid the explicit computation of features by developing methods that are able to compare structured data (e.g.,
sequences, graphs, trees). Even though feature-based methods for RE work well \cite{Jiang07asystematic},
there has been an increasing interest in exploiting kernel-based methods, due to the fact that sentences are
better described as structures (e.g., sequences of words, parsing trees, dependency graphs).

In this paper, we describe a new supervised approach to RE that is based on labeled dependency graph representations of the
sentences. The advantage is that a representation of a sentence as a labeled dependency
graph contains rich semantic information that, typically, contains useful hints when discriminating whether a set of entities
in a sentence are related. The solution we propose uses kernels to deal with these structures. We propose
the application of a marginalized kernel to compare labeled graphs \cite{Kashima03marginalizedkernels}. This kernel is based on
walks on random graphs and is able to exploit an infinite dimensional feature space by reducing its computation to the problem of
solving a system of linear equations. In order to make this graph kernel suitable for RE, we modified the kernel
to exploit the following properties that were previously introduced proposals of kernels for RE: \textit{(i)} the
words between the candidate entities or connecting them in a syntactic representation are particularly likely to carry
information regarding the relationship \cite{bunescu:emnlp05}; and \textit{(ii)} combining information from distinct sources in a 
kernel may help the RE system make better decisions \cite{Giuliano06exploitingshallow}.

In order to evaluate the model we propose, we performed some experiments with a biomedical dataset called AImed \cite{bunescu2006subsequence}.
This dataset is composed of several abstracts from Biology papers. The documents are annotated with interaction
relationships between proteins. The results show that the performance of our approach is comparable to the state-of-the-art.
Morever, when combining our kernel with other kernel methods, we were able to outperfom other state-of-the-art kernel methods.

The rest of the paper is organized as follows. In Section \ref{related}, we present the related work. 
Section \ref{probDefinition} defines the problem that we are trying to solve. In Section \ref{method}, we describe our
method for relationship extraction. In Section \ref{exp}, we report on the experiments performed.
Finally, Section \ref{conclusion} presents the conclusions and some topics for future work.

\section{Related Work}
\label{related}

The most relevant works in the topic of this paper are the ones that propose kernel methods for RE.
In the past ten years, several autors proposed kernels for different syntactic and semantic structures of a sentence.
One of the first approaches, presented in 2003 by Zelenko et al. \cite{Zelenko03kernelmethods}, is a kernel based
on shallow parse tree representation of sentences. This approach had some problems in what concerns the vulnerability to
parsing errors. In order to overcome these problems, Culotta and Sorensen \cite{1219009} proposed a generalization of this
kernel that, when combined with a bag-of-words kernel, is able to compensate the parsing errors.

In 2005, Bunescu and Mooney \cite{bunescu:emnlp05} proposed a kernel based on the shortest path between entities
in a dependency graph. The kernel was based on the hypothesis that the words between the candidate entities or connecting
them in a syntactic representation are particularly likely to carry information regarding the relationship. The problem of
this kernel is the fact that it is not very flexible when comparing candidates, which leads to very low values
of recall when the training data is too small. The same authors proposed a different kernel based on subsequences
\cite{bunescu2006subsequence}. The subsequences used in this approach could be combinations of words and other tags (e.g.,
POS tags, Wordnet Synsets). The results of this kernel are very interesting and even today it is still pointed out as a kernel
with a very good performance in RE tasks.

Giuliano et al. \cite{Giuliano06exploitingshallow} proposed in 2006 a kernel based only on shallow linguistic information of the sentences. The idea was to
exploit two simple kernels that, when combined, were able to obtain very interesting results. The \textit{global context
kernel} compares the whole sentence using a bag-of-n-grams approach. The frequencies of the n-grams are computed in three different
locations of the sentence: \textit{(i)} before the first entity; \textit{(ii)} between the two entities; and \textit{(iii)}
after the second entity. The \textit{local context kernel} evaluates the similarity between the entities of the sentences 
as well as the words in a window of limited size around them. The advantage of this kernel is its simplicity since it does not
need deep Natural Language Processing tools to preprocess the sentences in order to compute the kernel. However, its major advantage
may very well be a big disadvantage since it is not able to exploit rich syntactic/semantic information like a parsing tree or a
dependency graph representation of a sentence (which are structures that can be useful for determining whether a set of entities are
related).

In 2008, Airola et al. \cite{citeulike:3115793} presented a kernel that combines two graph representations of a sentence:
\textit{(i)} a labeled dependency graph; and \textit{(ii)} a linear order representation of the sentence. The kernel
considers all possible paths connecting any two vertices in the graph. The results obtained are comparable with the
state-of-the-art results. However, this kernel is very demanding in terms of computational resources.

In 2010, Tikk et al. \cite{PPIBenchmark} performed a study to analyze how a very comprehensive set of kernels for relationship
extraction performs when dealing the task of extracting protein-protein interactions. Even though they were not able to determine a
clear winner in their comparison, they were still able to outline some very interesting conclusions. First, they notice that kernels
based on dependency parsing tend to obtain better results than kernels based on tree parses. Moreover, they show that a simple
kernel, like \cite{Giuliano06exploitingshallow}, can still obtain results that are at the level of the best kernels based on
dependency parsing.

\section{Problem Definition}
\label{probDefinition}

In general, the problem of finding an \textit{n}-ary relationship between entities can be seen as a classification problem for
which the input is a set of \textit{n} entities and the output is the type of relationship between them or an indication that
they are not related at all.

With this definition, given a text document with all the entities identified, the candidate results are all the sets
of $n$ entities that exist in the text. This approach would generate a huge set of candidates among which very
few correspond to actually related entities. For this reason, this configuration would potentially lead to some performance
issues (due to the huge amount of candidates) and to some problems in terms of accuracy (due to the unbalancement
of the data). To avoid these issues, we exploit an heuristic that is typically used in related works, which consists in limiting 
the candidates to sets of entities that can be found in the same sentence.

This way, for one sentence with $k$ entities, the number of candidates generated for a $n$-ary
relationship is given by the number of combinations of the $k$ entities, selected $n$ at a time, i.e. 
$\binom{k}{n}$. For instance, consider the sentence in Figure \ref{fig:sentExample}, in which we present an example of a
sentence from a biomedical text. Suppose that we aim at finding interaction relationships between proteins. This sentence 
contains three identified proteins: \textit{TRADD}, \textit{RIP} and \textit{Fas}. Moreover, there are two interaction relationships
between these entities: \textit{TRADD} interacts with \textit{RIP} and \textit{RIP} interacts with \textit{Fas}.

\begin{figure}[tb]
    \centering
    \includegraphics[scale=0.30]{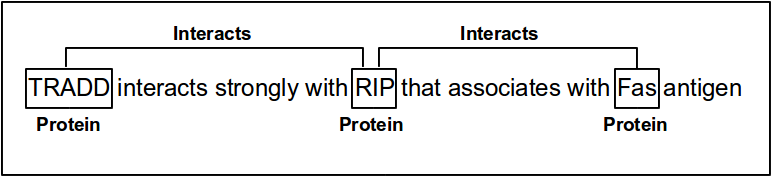}
    \caption{A sentence from a biomedical text containing three references to proteins (\textit{TRADD}, \textit{RIP} and \textit{Fas})
    and two interaction relationships between them (\textit{TRADD} interacts with \textit{RIP} and \textit{RIP} interacts with \textit{Fas}).}
    \label{fig:sentExample}
\end{figure}

Given the fact that a protein interaction is a binary relationship, we have a total of $\binom{3}{2}=3$ candidates, which are
presented in Figure \ref{fig:sentExample2}.

\begin{figure}[tb]
    \centering
    \includegraphics[scale=0.30]{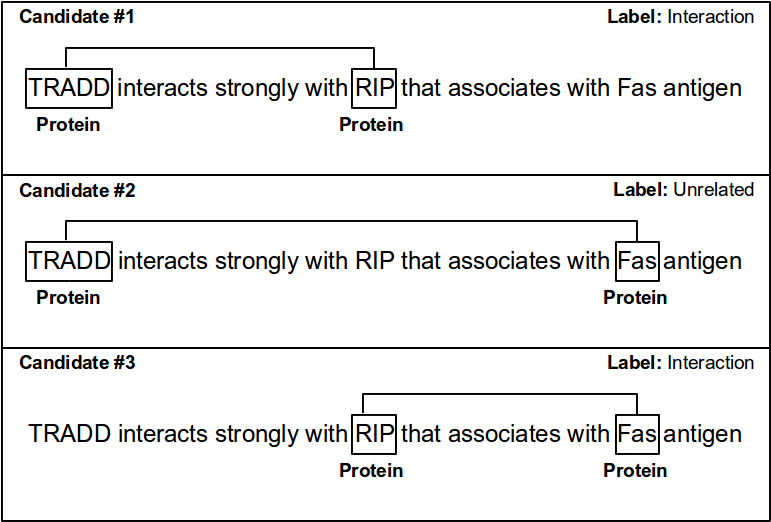}
    \caption{Candidates generated from the sentence of Figure \ref{fig:sentExample}.}
    \label{fig:sentExample2}
\end{figure}

Note that it is also possible to use other heuristics to reduce the number of candidates. For instance, in some cases, we may
have knowledge about the types of entities that can fulfill a given role in a relationship (e.g. in a relationship between a
company and its CEO, it is known that one of the entities must be a a company and the other, a person). Even though these heuristics
typically involve some type of prior knowledge about the application domain, they tend to drastically reduce the space of
candidates. This fact makes the relationship extraction process a lot easier and helps it produce better results since some of the
candidates involving entities that are never related are not used.

Assuming a set of candidate results, Figure \ref{fig:workflow} describes how the RE extraction task can be represented as a
classification problem. The problem can be divided into two main phases: training and execution. In the
\textit{training phase}, the objective is to automatically generate a \textit{statistical model} that is able to determine
whether a given candidate  corresponds to a relationship. In order to produce this model, some training examples must be provided
to a \textit{learning algorithm} (e.g., solving a quadratic optimization problem in the case of a SVM classifier). These examples
are generated in the same fashion as the candidates, however, they include an additional label that indicates whether they
correspond to a relationship.

\begin{figure}[t]
    \centering
    \includegraphics[scale=0.27]{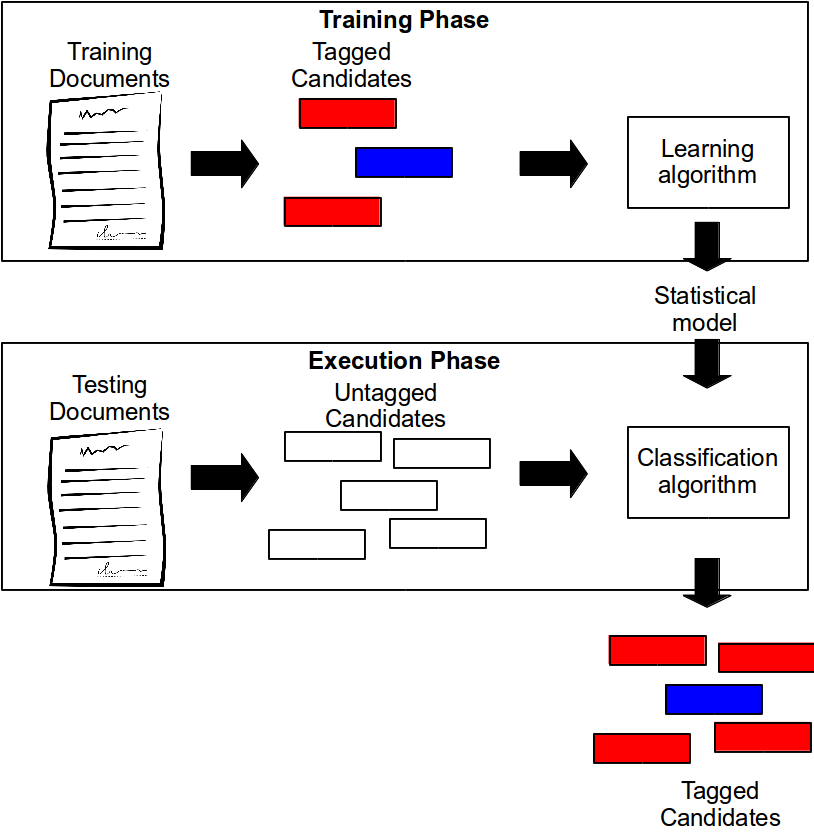}
    \caption{Representation of a RE task as a classification problem.}
    \label{fig:workflow}
\end{figure}

The \textit{execution phase} aims at classifying each unlabeled candidate from new untagged documents as containing a relationship
or not. This decision is made using the statistical model created in the training phase and a \textit{classification algorithm}. In
the end of the process, the sets of entities in the candidates that are classified as containing a relationship are returned.

\section{Method}
\label{method}

In this Section, we present the proposed kernel method. We start by describing the basic idea
behind kernel methods for RE in Section \ref{kernelMethods}. Then, in Section \ref{graphRep}, we propose a representation 
of the candidate sentences as labeled graphs. In Section \ref{rwKernel}, we explain the random walk kernel that was used
as the basis for our RE kernel. In Section \ref{paramsKernel}, we present the parameters used to modify the random
walk kernel for our problem. Finally, in Section \ref{kernel}, we propose our kernel for RE.

\subsection{Kernel Methods for Relationship Extraction}
\label{kernelMethods}

In some cases, input objects of a classifier may not be easily expressed via feature vectors (e.g., if the range of possible
features is too wide or if the nature of the object does not make it clear how to choose the features). Therefore, the feature
engineering process may become painfully hard and lead to high-dimensional feature spaces and consequently to computational
problems. Kernel methods are an alternative to feature-based methods that can be used to classify objects while keeping their
original representation.

In kernel methods, the idea is to exploit a similarity function (kernel) between input objects. This function, with the
help of a discriminative machine learning technique, is used to classify new examples. In order for a similarity
function to be an acceptable kernel function, $K(x,y)$, it must respect the following properties: \textit{(i)} it must be a
bidimentional function over the object space \textit{X} to a number in
$[0, +\infty[$ ($K:$ $X$ $\times$ $X$ $\xrightarrow{}$ $[0, +\infty[$);
\textit{(ii)} it must be symmetric ($\forall_{x,y\epsilon X},K(x,y)=K(y,x)$); and
\textit{(iii)} it must be positive-semidefinite ($\forall_{x_1,x_2,...x_n \epsilon X},the$ $n \times n$ matrix
$(K(x_i,x_j))_{ij}$ is positive-semidefinite).

RE is an example of a problem for which the inputs may not be easily expressed via feature vectors. As described in
Section \ref{probDefinition}, the inputs of the learning and classification algorithms in supervised RE tasks are sentences.
Typically, sentences are better described as structures (e.g., sequences of words, parsing trees, dependency graphs) and 
it is interesting to use these representations directly.

\subsection{Labeled Graph Representation of the Sentences}
\label{graphRep}

In our approach, we assume that the inputs of the learning and classification algorithms are labeled graph representations 
of the candidate sentences (see Figure \ref{fig:graph}). In this graph, each vertex is associated with a word in the sentence
and is enriched with additional features of the word. In our representation, the additional features include POS tags, generic
POS tags, the lemma of the word and capitalization patterns (however, due to simplicity, we represent only one additional feature
in the graph of Figure \ref{fig:graph} which is the POS tag). We could use other potentially useful features like hypernyms or
synsets extracted from the WordNet. The edges represent semantic relationships between the words. The type of the semantic 
relationship is represented by the edge label.

\begin{figure*}[t]
    \centering
    \includegraphics[scale=0.3]{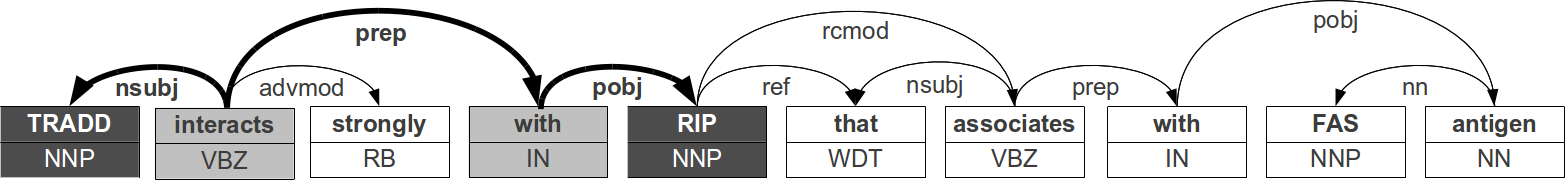}
    \caption{Graph Representation of Candidate \#1 presented in Figure \ref{fig:sentExample2}. Each node is composed by the word
    and its POS tag. The candidate entities are represented in black. We also represent the shortest path between the two entities
    with dark edges. The nodes that cross the shortest path are represented in gray.}
    \label{fig:graph}
\end{figure*}

Recall that, for a given sentence with \textit{k} entities, when searching for a n-ary relationship, the number of candidates
that are generated is $\binom{k}{n}$. In terms of structure (vertexes and edges), the corresponding dependency graph
for each of these candidates is always the same. If we used only structural information to compare candidates we
could have a problem because we would not be able to distinguish between different candidates generated from the
same sentence that are expected to produce different classification results.

For this reason, we used heuristics to enrich our graph representation. First, the entities that are candidate
to be related can provide very important clues for detecting if there is a relationship \cite{Giuliano06exploitingshallow}.
We define a predicate $isEntity(v)$, which receives a vertex of the graph and determines whether it is an entity.
With this, it is possible for a kernel to use this information in the computation of the similarity between graphs.
Second, the shortest path hypothesis, formalized in \cite{bunescu:emnlp05}, states that
the words between the candidate entities or  connecting them in a syntactic representation are particularly likely to carry
information regarding their relationship. Analogously to \cite{bunescu:emnlp05} and \cite{citeulike:3115793}, we exploited
this hypothesis by defining a predicate called $inSP(x)$ that receives as input a node or an edge of the graph and returns
true if they belong to the shortest path between the two entities of the graph. Like in the case of the entities, this allows the
kernel to treat these vertexes and edges in a special fashion way.

\subsection{Random Walks Kernel}
\label{rwKernel}

The random walk kernel used as a basis of our RE kernel was defined in \cite{Kashima03marginalizedkernels} as a
marginalized kernel between labeled graphs. The basic idea behind this kernel is the following one: given a pair of graphs,
perform simultaneous random walks between the vertexes of the graphs and count the number of matching paths. In a more formal way,
the objective of the kernel is to compute the expected number of matching paths between the two graphs.

In order to explain this kernel, we start by defining the graph that is expected as input. Let $G$ be a labeled
directed graph and $|G|$ be the number of vertexes in the graph. All vertexes in the graph are labeled and $v_i$ denotes
the label of vertex $i$. The edges of the graph are also labeled and $e_{ij}$ denotes the label of
the edge that connects vertex $i$ and vertex $j$. Moreover, we assume two kernel functions,
$K_v(v,v')$ and $K_e(e,e')$ that are kernel functions between vertexes and edges respectively. Figure
\ref{fig:graphKernelExample} presents an example of a graph that can be used as input of the random walk kernel.

\begin{figure}[t]
    \centering
    \includegraphics[scale=0.45]{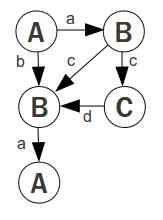}
    \caption{Example of a labeled graph that can be used as input of the Random walk kernel.}
    \label{fig:graphKernelExample}
\end{figure}

Additionally to the graph, this kernel also assumes the existence of three probability distributions:
\textit{(i)} the initial probability distribution, $p_s(h)$, that corresponds to the probability that a path starts
in the vertex $h$; \textit{(ii)} the ending probability, $p_q(h)$, that corresponds to the probability that a path ends
in the vertex $h$; and \textit{(iii)} the transition probability, $p_t(h_i|h_{i-1})$, that corresponds to the
probability that we walk from vertex $h_{i-1}$ to vertex $h_{i}$. With all these probabilities defined, it is possible
to compute the probability of a path $\textbf{h}=[h_1,h_2,..., h_l]$ in the graph $G$ with Equation \ref{eq:probPath}.

\begin{equation}
    \centering
    \scriptsize
    p(\textbf{h}|G)=p_s(h_1)\prod_{i=2}^lp_t(h_i|h_{i-1})p_q(h_l)
    \label{eq:probPath}
\end{equation}

As we stated before, the objective of the kernel is to compute the expected number of matching paths between two input
graphs. Let us define a kernel to compute the number of matching subpaths between two paths of different graphs. We assume
that if the paths have different lenghts, then there is no match between them. If the paths have the same length, the matching
between them is given by the product of the vertex and edge kernels.
Assuming we have two paths $\textbf{h}$ and $\textbf{h'}$ from two different graphs $G$ and $G'$, then the kernel
between $z=(\textbf{h},G)$ and $z'=(\textbf{h'},G')$ is given by Equation \ref{eq:kernelPaths}.

\begin{equation}
    \centering
    \scriptsize
    K_z(z,z')=\left\{ 
		  \begin{array}{l l}
		    0 & \quad \text{if $l \neq l'$}\\
		    K_v(v_{h_1},v_{h'_1}')\prod_{i=2}^lK_v(v_{h_i},v_{h'_i}')\times & \quad \text{if $l = l'$}\\
			   \hspace{2cm} K(e_{h_{i-1}h_{i}},e_{h_{i-1}'h_{i}'}') \\
		  \end{array} \right.
    \label{eq:kernelPaths}
\end{equation}

Given $K_z(z,z')$ and $p(\textbf{h}|G)$, we can compute the expected number of matching paths between the two graphs with
Equation \ref{eq:kernelGraphs}.

\begin{equation}
    \centering
    \scriptsize
    K(G,G')=\E[K_z(z,z')]=\sum_{\textbf{h}}\sum_{\textbf{h'}}K_z(z,z')p(\textbf{h}|G)p(\textbf{h'}|G')
    \label{eq:kernelGraphs}
\end{equation}

Computing this kernel using a naive approach (i.e., going through all the possible pairs of paths in the kernels), would be
computational expensive for acyclic graphs and impossible for graphs containing cycles. However,
\cite{Kashima03marginalizedkernels} demonstrated that this kernel can be efficiently computed by
solving a system of linear equations. In order to define this system of linear equations, let us first define the following
matrices:

\[
 \scriptsize
 S =
 \begin{pmatrix}
  \scriptsize
  s(1,1') \\
  s(1,2') \\
  \vdots  \\
  s(1,|G'|') \\
  s(2,1') \\
  \vdots  \\
  s(|G|,|G'|')
 \end{pmatrix}
 \hspace{0.2cm}
  Q =
\begin{pmatrix}
  q(1,1') \\
  q(1,2') \\
  \vdots  \\
  q(1,|G'|') \\
  q(2,1') \\
  \vdots  \\
  q(|G|,|G'|')
 \end{pmatrix}
\]

\[
\scriptsize
 T \hspace{-0.1cm}=
 \hspace{-0.1cm}
 \begin{pmatrix}
  t(1,1',1,1') & \hspace{-0.2cm} t(1,1',1,2') &\hspace{-0.3cm} \cdots &\hspace{-0.3cm} t(1,1',|G|,|G'|') \\
  t(1,2',1,1') & \hspace{-0.2cm} t(1,2',1,2') &\hspace{-0.3cm} \cdots &\hspace{-0.3cm} t(1,2',|G|,|G'|') \\
  \vdots  & \hspace{-0.2cm} \vdots  & \hspace{-0.3cm} \ddots &\hspace{-0.3cm} \vdots  \\
  t(|G|,|G'|',1,1') & \hspace{-0.2cm} t(|G|,|G'|',1,2') & \hspace{-0.3cm} \cdots &\hspace{-0.3cm} t(|G|,|G'|',|G|,|G'|') \\
 \end{pmatrix}
\]

Where

\begin{equation}
 \scriptsize
 s(h_1,h_1')=p_s(h_1)p_{s'}(h_1')K_v(v_{h_1},v_{h_1'}')
\end{equation}

\begin{equation}
 \scriptsize
 q(h_l,h_l')=p_q(h_l)p_{q'}(h_l')
\end{equation}

\begin{equation}
\scriptsize
\begin{array}{l}
 t(h_{i-1},h_{i-1}',h_i,h_i') = p_t(h_i|h_{i-1})p_t(h_i'|h_{i-1}') \times \\
    \hspace{2cm}K_v(v_{h_i},v_{h'_i}')K(e_{h_{i-1}h_{i}},e_{h_{i-1}'h_{i}'}')
\end{array}
\end{equation}

The system of linear equations that we need to solve is presented in Equation \ref{eq:linearSystem}

\begin{equation}
 \scriptsize
 (I-T)X=Q
 \label{eq:linearSystem}
\end{equation}

where X is the solution of the system and I is the identity matrix. \cite{Kashima03marginalizedkernels} demonstrated that the
random walk kernel between graphs, $K(G,G')$, can be given by Equation \ref{eq:kernelSimple}.

\begin{equation}
    \scriptsize
    \centering
    K(G,G')=<S,X>
    \label{eq:kernelSimple}
\end{equation}

where $<S,X>$ is the inner product between two vectors.

\subsection{Parameters of the Random Walks Kernel for Relationship Extraction}
\label{paramsKernel}

In Section \ref{rwKernel}, we described a kernel for generic labeled graphs. The kernel we propose is a particularization
of this one applied to RE.

Recall that our representation of a sentence, presented in Section \ref{graphRep} corresponds to
a labeled graph where the labels of the vertexes are vectors of tags (containing the word itself, its lemma, POS tags, 
and ortographic patterns) and the labels of the edges contain simply the type of the semantic relationship between the two
entities. Moreover, each vertex and edge contains information about whether it is in the shortest path between the
two entities. The vertexes also contain information about whether they are entities.

In order to use the random walk kernel described in Section \ref{rwKernel}, we had to define the kernels between the
vertex labels and the kernels between the edge labels. Given the fact that the labels of the vertexes are simply vectors
of attributes of the word associated with the vertex, we can use the normalized linear kernel presented in Equation
\ref{eq:normalizedKernel}.

\begin{equation}
    \centering
    \scriptsize
    K_v(v,v')=\frac{c(v,v')}{\sqrt{c(v,v)c(v',v')}}
    \label{eq:normalizedKernel}
\end{equation}

where $c(v,v')$ counts the number of common features between the labels of $v$ and $v'$.

In order to guarantee that entities can only match in a random walk with other entities and that vertexes
contained in a shortest path can only match with vertexes contained in a shortest path, we actually used a slightly
modified version of the kernel presented in Equation \ref{eq:normalizedKernel}. The modified version is presented in Equation
\ref{eq:normalizedKernel2}.

\begin{equation}
\label{eq:normalizedKernel2}
    \centering
    \scriptsize
    K_v(v,v')=\left\{ 
		  \begin{array}{l l l}
		    \frac{c(v,v')}{\sqrt{c(v,v)c(v',v')}} & \text{if} & \text{$inSP(v)=inSP(v')$ $\wedge$}\\
		                                          &  & \text{$isEntity(v)=isEntity(v')$} \\
		    & \\
		    0 & & \text{otherwise} \\
		  \end{array} \right.
\end{equation}

The kernel between the edges is very simple. Since the label for the edges is only a string indicating the type
of semantic relationship between the two words. We define this kernel in Equation \ref{eq:normalizedKernel3}.

\begin{equation}
    \centering
    \scriptsize
    K_e(e,e')=\delta(e=e')
    \label{eq:normalizedKernel3}
\end{equation}

where, $\delta$ is a function that returns 1 if its argument holds and 0 otherwise.

Once again, since we want to differenciate edges in the shortest path from edges outside the shortest path, we added a simple
modification to the kernel that is presented in Equation \ref{eq:normalizedKernel4}.

\begin{equation}
    \label{eq:normalizedKernel4}
    \centering
    \scriptsize
    K_e(e,e')=\left\{ 
		  \begin{array}{l l l }
		    \delta(e=e') & \text{if} & \text{$inSP(e)=inSP(e')$}\\
		    0 & & \text{otherwise}\\
		  \end{array} \right.
\end{equation}

Finally, we still need to define the probability distributions necessary to compute the random walk kernel in
our problem. Due to the fact that we have no prior knowledge about the probability distributions, we
follow the solution proposed in \cite{Kashima03marginalizedkernels} and consider that all the distributions are
uniform.

\subsection{Random Walks Kernel for Relationship \\Extraction}
\label{kernel}

Using the random walk kernel presented in Section \ref{rwKernel} and the parameterization for the RE problem proposed
in Section \ref{paramsKernel}, we produced three variations of the kernel: \textit{(i)} Full Graph Kernel; 
\textit{(ii)} Shortest Path Kernel; and \textit{(iii)} No Shortest Path Kernel.

The \textit{Full Graph Kernel} (FGK) corresponds to the application of the random walk kernel to the whole structure described in 
Section \ref{graphRep}. The idea of this kernel is to capture the whole view of the graph structure (which is the same for 
all the candidates generated from a given sentence) but still be able to capture the similarity between interesting
properties that are specific to the candidates (i.e., shortest path and entities information).

The \textit{Shortest Path Kernel} (SPK) aims at exploiting the shortest path hypothesis presented in \cite{bunescu:emnlp05}.
The idea is to apply the random walk kernel to the subgraph that corresponds to the shortest path between the entities.

The \textit{No Shortest Path Kernel} (NSPK) is a variation of FGK where the nodes and edges that belong to the
shortest path are not marked as such. For this reason, the only thing that distinguishes the graph structures for
candidates generated from a given sentence are the entities.

The kernel we propose is actually based on a very interesting property of kernels: the linear combination of several
kernels is itself a kernel. We used this approach because several works empirically demonstrated that combining kernels
using this approach typically improves the performance of individual kernels \cite{1219009,Giuliano06exploitingshallow}.

\section{Experiments}
\label{exp}

In this Section, we present the experiments performed in order to evaluate our solution for RE and report on
the results obtained. First, we present the relationship extraction task. Then,
in Section \ref{data}, we describe the dataset. Section \ref{metrics} presents the metrics used to evaluate our
kernel and Section \ref{significance} presents the method used to support our claims in what concerns the comparison
of the kernels. In Section \ref{implem}, we point out some implementation
details of our experiments. In Section \ref{indiv} we report on the performance of the individual kernels presented in
Section \ref{kernel} and in Section \ref{combo} we report on the combination of these kernels. In Section \ref{comp}, we
perform a comparison between our solution and other methods. Finally, in Section \ref{comboplus} we report on some
experiments when combining our kernel with other methods.

\subsection{Relationship Extraction Task}
\label{task}

In our evaluation, we focused exclusively on the extraction of relationships that correspond to protein-protein interactions.
The idea is that, given pairs of entities there is a relationship between them if the text indicates that the proteins
have some kind of biological interaction.

\subsection{Dataset}
\label{data}

We performed our experiments over a protein-protein interaction dataset called AImed\footnote{ftp://ftp.cs.utexas.edu/pub/mooney/bio-data/interactions.tar.gz}. 
This dataset has been used in previous works to evaluate the performance of relationship extraction systems in the task of
extracting protein-protein interactions \cite{bunescu2006subsequence,Giuliano06exploitingshallow,citeulike:3115793}. AImed is
composed by 225 Medline abstracts from which 200 describe interactions between proteins and the other 25 do not refer to any
interaction. The total number of interacting pairs is 974 and the total number of non-interacting pairs is 4072.

During the evaluation of our model we used a cross-validation strategy that is based on splits of the AImed dataset
at the level of document \cite{bunescu2006subsequence,citeulike:3115793}. Table \ref{table:splits} presents the number of
positive and negative candidates that can be found in the training and testing data of each split.

\begin{table}[t]
  \scriptsize
  \centering
  \begin{tabular}{| c | c | c | c | c | }
    \hline  
    split & \# Pos Train & \# Neg Train & \# Pos Test & \# Neg Test\\
\hline 
\hline                       
    1 & 866 & 3675 & 108 & 397 \\
    2 & 896 & 3813 & 78 & 259 \\    
    3 & 894 & 3626 & 80 & 446 \\
    4 & 872 & 3395 & 102 & 677 \\
    5 & 865 & 3731 & 109 & 341 \\
    6 & 854 & 3563 & 120 & 509 \\
    7 & 876 & 3735 & 98 & 337 \\
    8 & 883 & 3765 & 91 & 307 \\
    9 & 894 & 3718 & 80 & 354 \\
    10 & 866 & 3627 & 108 & 445 \\
    \hline  
  \end{tabular}
  \caption{Number of training and testing candidates for each split}
  \label{table:splits}
\end{table}

\subsection{Evaluation Metrics}
\label{metrics}

Our experiments are focused on measuring the quality of the results produced when using our kernel. In Information Extraction
(and particularly in Relationship Extraction), the quality of the results produced is based on two metrics: recall and precision.

\emph{Recall} gives the ratio between the amount of information correctly extracted from the texts and the information available
in texts. Thus, recall measures the amount of relevant information extracted and is given by Equation \ref{label:Recall}:

\begin{equation}
    \scriptsize
    recall=\frac{C}{P}
    \label{label:Recall}
\end{equation}

where $C$ represents the number of correctly extracted relationships while $P$ represents the total number of relationships that
should be extracted. The disadvantage of this measure is the fact that it returns high values when we extract all possible pairs
of entities as a relationship regardless of them being related or not.

\emph{Precision} is the ratio between the amount of information correctly extracted from the texts and all the information
extracted. The precision is then a measure of confidence on the information extracted and is given by
Equation \ref{label:Precision}:

\begin{equation}
    \scriptsize
    precision=\frac{C}{C+I}
    \label{label:Precision}
\end{equation}

where $C$ represents the number of relationships correctly extracted, $I$ represents the number of relationships incorrectly
extracted.

The disadvantage of precision is that we can get high results extracting only information that we are sure to be right and
ignoring information that are in the text and may be relevant.

The values of recall and precision may enter in conflict. When we try to increase the recall, the value of precision may decrease
and vice versa. The \emph{F-measure} was adopted to measure the general performance of a system, balancing the values of recall
and precision. It is given by Equation \ref{label:MedidaF}:

\begin{equation}
    \scriptsize
    F\text{-}measure=\frac{(\beta^{2}+1)\times P\times R}{\beta^{2}\times P + R}
    \label{label:MedidaF}
\end{equation}

where $R$ represents the recall, $P$ represents the precision, \begin{math}\beta\end{math} is an adaptation value of the equation
that allows to define the relative weight of recall and precision. The value \begin{math}\beta\end{math} can be interpreted as the
number of times that the recall is more important than accuracy. A value for \begin{math}\beta\end{math} that is often used is 1, in
order to give the same weight to recall and precision. In this case, the \emph{F-measure} value is obtained through Equation
\ref{label:MedidaFbeta1}:

\begin{equation}
    \scriptsize
    F_1=\frac{2\times P\times R}{P+R}
    \label{label:MedidaFbeta1}
\end{equation}

\subsection{Significance Tests}
\label{significance}

In order to support our claims during the comparison of each pair of kernels, we relied significance tests. We used a
the paired t-test between each pair of kernels that we wanted to compare directly. Details about
this significance test can be found on most statistics text books \cite{statisticsForDummies}.

For a given metric presented in Section \ref{metrics}, we give as input to the test the result obtained for each split of the
dataset. Our claims are based on a significance level of 5\%.

\subsection{Implementation Details}
\label{implem}

Our experiments used the SVM package jLIBSVM\footnote{http://dev.davidsoergel.com/trac/jlibsvm/}, 
a Java port of LIBSVM that allows for easy customization when using different kernels. During the experiments, we
used most of the default parameters of jLIBSVM. The only exception was the parameter C of the SVM (which controls the
trade-off between the errors of the SVM and the size of the margin). For this parameter, after some
empirical experimentation we fixed its value in 50 for all the experiments.

We used the OpenNLP\footnote{http://incubator.apache.org/opennlp/} module for 
sentence detection and the Stanford parser\footnote{http://nlp.stanford.edu/software/lex-parser.shtml} for the word 
segmentation, POS tagging and generation of the labeled dependency graph.

Finally, we used Parallel Colt\footnote{http://sites.google.com/site/piotrwendykier/software/} to perform the
matrix operations necessary for our kernel.

\subsection{Performance of the Individual Kernels}
\label{indiv}

Our first experiment aimed at understanding how each of the individual kernels that we proposed (i.e., $FGK$, $SPK$ and $NSPK$ introduced
in Section \ref{kernel}) performs. Table \ref{table:indiv} shows the results of this experiment.

\begin{table}[t]
  \centering
  \begin{tabular}{| c | c | c | c |}
    \hline  
    Kernel & Recall & Precision & $F_1$ \\
\hline 
\hline                       
    $FGK$ & 41.51\% & \textbf{58.94\%} & 48.25\% \\ 
    $SPK$ & \textbf{43.47\%} & 56.73\% & \textbf{48.86\%} \\   
    $NSPK$ & 37.69\% & 58.47\% & 45.39\% \\
    \hline
  \end{tabular}
  \caption{Performance of the individual kernels on the AImed data set.}
  \label{table:indiv}
\end{table}

The results obtained are according to what was expected. First, the individual 
kernel that obtains the highest value of $F_1$ is $SPK$. Knowing how the shortest path hypothesis has been
exploited with success in several other works, this comes with no surprise. Even though the average value of $F_1$
for $SPK$ is higher than that for $FGK$, the difference is not statistically significant according to the significance tests.

If we look only at the average values of recall and precision presented in Table \ref{table:indiv}, it seems
that $SPK$ is the best kernel in terms of recall and $FGK$ is the best in terms of precision. However, by comparing
the results obtained by these two kernels using the significance tests the differences are not significant for both these
metrics.

Another result that is not surprising is the fact that the performance of $NSPK$ is very poor. As discussed before,
this kernel does not distinguish very well candidates that are generated from the same sentence but are associated with
different pairs of entities. This reflects in a drastic drop of the recall value.

\subsection{Performance of the Combination of Kernels}
\label{combo}

After analyzing the performance of the individual kernels, we evaluated the performance
of the kernels that result from their combination. We considered the
following four combinations: \textit{(i)} $FGK+SPK$; \textit{(ii)} $FGK+NSPK$; \textit{(iv)} $SPK+NSPK$;
and \textit{(iii)} $ALL=FGK+SPK+NSPK$.

Table \ref{table:comb} shows the results of this experiment. Given the performance of the individual
kernels reported before, it was expected that the best combination of kernels would be either the one that
combines all the individual kernels ($ALL$) or the one that combines the two best individual kernels ($FGK+SPK$).
In fact, the results show that regarding the average values of recall, precision and $F_1$, the best combination is
actually $SPK+NSPK$. 

The explanation for this surprising result has to do with the definition of these kernels. On one side, $SPK$ was
designed as a good solution to distinguish between candidates generated from the same sentence and associated with
different pairs of entities. On the other side, $NSPK$ is good to analyze the whole structure of the dependency
graph but it does not distinguish very well the candidates generated from the same sentence. Thus, these two kernels
are good at distinguishing very different contexts of the candidates. For this reason, they end up being a good
complement to each other.

Even though $SPK+NSPK$ obtained the best average values of recall, precision and $F_1$, it is important to note
that according to the significance tests, it is not fair to claim that it is a superior solution in comparison to
$FGK+SPK$ and $ALL$ since the differences for all the metrics were not statistically significant.

\begin{table}[t]
  \centering
  \begin{tabular}{| c | c | c | c |}
    \hline  
    Kernel & Recall & Precision & $F_1$ \\
\hline 
\hline                      
    $FGK+SPK$ & 45.21\% & 59.60\% & 51.83\% \\ 
    $FGK+NSPK$ & 40.84\% & 57.56\% & 47.34\% \\ 
    $SPK+NSPK$ & \textbf{46.41\%} & \textbf{60.57\%} & \textbf{52.31\%} \\ 
    $ALL$ & 46.31\% & 59.01\% & 51.64\% \\ 
    \hline
  \end{tabular}
  \caption{Performance of the individual kernels on the AImed data set.}
  \label{table:comb}
\end{table}

Another interesting observation has to do with the terrible results obtained by $FGK+NSPK$. It is the kernel combination
with worst results in all the metrics. Moreover, the significance tests indicated that in terms of recall and $F_1$ measure,
the differences in comparison to the other combinations were significant. These results are also related with the type of
information that the two individual kernels try to analyze. Recall that $FGK$
is actually a modified and more refined version of $NSPK$ in which vertexes and edges of the shortest path between
the candidate entities are treated differently. For this reason, most of the information exploited by both kernels is the
same, which makes their combination a little bit redundant.

Finally, we wanted to compare the combination kernels with the individual kernels to understand whether it
pays off to use the combinations. For each metric, we compared the combination kernels with the individual kernel with the highest
value of the metric as presented in Table \ref{table:indiv}. First, in what concerns recall, we observe that the differences
between $SPK$ and most of the combinations is not significant. The only exception is $SPK+NSPK$. In what concerns precision,
we compared with $FGK$ and we observed that the gains from using the combinations in this case are not significant. For, the
comparison regarding $F_1$, most of the combination kernels significantly outperform $SPK$. The only exception is $FGK+NSPK$.
In fact, if we compare $FGK+NSPK$ with both kernels that originate it, we notice that the differences in terms of $F_1$ between
them are not statistically significant. This is interesting because it illustrates how combining two kernels does not necessarily
mean that the results will improve.

\vspace{1cm}

\subsection{Comparison with Other Methods}
\label{comp}

In order to compare the performance of our solution with other methods, we implemented two additional kernels described
in the literature: \textit{(i)} a kernel based on shallow linguistic information of the sentences, \cite{Giuliano06exploitingshallow};
and \textit{(ii)} a kernel based on subsequences, \cite{bunescu2006subsequence}. During these experiments we always compared these
kernels with our combination of kernels that showed better performance on the average values of the recall, precision and $F_1$:
$SPK+NSPK$. Table \ref{table:comp} shows the results of this experiment.

\begin{table}[t]
  \centering
  \begin{tabular}{| c | c | c | c |}
    \hline
    Kernel & Recall & Precision & $F_1$ \\
\hline 
\hline  
    \cite{Giuliano06exploitingshallow} & \textbf{47.74\%} & 62.09\% & \textbf{53.49\%} \\ 
    \cite{bunescu2006subsequence} & 41.15\% & \textbf{66.68\%} & 50.60\% \\ 
    $SPK+NSPK$ & 46.41\% & 60.57\% & 52.31\% \\
    \hline
  \end{tabular}
  \caption{Performance of the individual kernels on the AImed data set}
  \label{table:comp}
\end{table}

The most evident conclusion obtained by observing the results is that our solution is still
outperformed by the shallow linguistic information kernel in terms of average values of the metrics.
However, the significance tests for all the metrics indicate that the differences between $SPK+NSPK$ and
\cite{Giuliano06exploitingshallow} are not significative.

If we compare $SPK+NSPK$ with \cite{bunescu2006subsequence}, the results are very different. In fact, the results
of the significance tests show that there are significant differences between these two kernels in terms of recall
and precision ($SPK+NSPK$ is better in terms of recall and \cite{bunescu2006subsequence} is better in terms of precision).
However, in terms of $F_1$, the differences are not significative (even though the $SPK+NSPK$ obtains an higher average value
of $F_1$).

The differences of the results of precision and recall of $SPK+NSPK$ and \cite{Giuliano06exploitingshallow} in comparison to
\cite{bunescu2006subsequence} are something worth mentioning: the precision values are not
as high as in the subsequences kernel but the values of recall are significantly higher. This is interesting because
it goes against a typical trend in works on supervised RE in which the values of precision tend to be very high
but the values of recall tend to be very low.

\subsection{Combination with Other Kernel Methods}
\label{comboplus}

Finally, we performed some experiments to evaluate how combining $SPK+NSPK$ with other methods influences
the results. Once again, we used the two kernels that we compared our solution to in Section \ref{comp}.
Table \ref{table:combplus} presents the results of this experiment.

\begin{table}[t]
  \centering
  \begin{tabular}{| c | c | c | c |}
    \hline  
    Kernel & Recall & Precision & $F_1$ \\
\hline 
\hline                     
    {\scriptsize $SPK+NSPK$ + \cite{Giuliano06exploitingshallow}} & \textbf{49.38\%} & 64.12\% & \textbf{55.43\%} \\ 
    {\scriptsize $SPK+NSPK$ + \cite{bunescu2006subsequence}} & 45.67\% & 67.96\% & 54.23\% \\ 
    {\scriptsize \cite{Giuliano06exploitingshallow} + \cite{bunescu2006subsequence}} & 45.21\% & \textbf{69.07\%} & 54.12\% \\ 
    {\scriptsize $SPK+NSPK$ + \cite{Giuliano06exploitingshallow} + \cite{bunescu2006subsequence}}  & 46.66\% & 68.36\% & 55.14\% \\ 
    \hline
  \end{tabular}
  \caption{Performance of the individual kernels on the AImed data set}
  \label{table:combplus}
\end{table}

By analyzing the results obtained in this experiment, we observe that the best combination is the one that joins
$SPK+NSPK$ with \cite{Giuliano06exploitingshallow}. Moreover, even the combination of $SPK+NSPK$ with \cite{bunescu2006subsequence}
is able to outperform the combination of \cite{Giuliano06exploitingshallow} and \cite{bunescu2006subsequence}.

In order to understand these results, recall that \cite{Giuliano06exploitingshallow}  is based on several kernels
including information of n-grams in three different locations of the sentence: before the first entity, between the entities
and after the second entity. Knowing that n-grams are among the subsequences of the sentence, it is easy to undestand
that there is some overlapped information when combining these two kernel.

When these kernels are combined with $SPK+NSPK$, we are joining information from completely different sources: sequences
and dependency graph. For this reason, the kernel we propose is very interesting when used in combinations with kernels
from different sources.

We also wanted to determine whether the difference of the results of these combinations in comparison to the individual kernels
was significative. Thus, we performed significance tests between $SPK+NSPK$, \cite{Giuliano06exploitingshallow}, \cite{bunescu2006subsequence}
and all their combinations presented in Table \ref{table:combplus}.

In what concerns recall, the differences between the combinations, $SPK+NSPK$ and \cite{Giuliano06exploitingshallow} are not
significative. However, the tests indicate that all the combinations are able to outperform \cite{bunescu2006subsequence}. This comes with no surprise
knowing that the differences in terms of average value of recall were very high.

Regarding precision, the significance tests show that combining $SPK+NSPK$ with all the other kernels have a significant impact.
The tests also obtain the same result for \cite{Giuliano06exploitingshallow}. With \cite{bunescu2006subsequence} the results are different: none of the
combinations is able to significantly outperform \cite{bunescu2006subsequence}.

When comparing the results of the significance tests for $F_1$, there is only one combination that is able to clearly outperforms
$SPK+NSPK$ and \cite{Giuliano06exploitingshallow}. This combination is actually the one that combines both these kernels. In all
the other cases, the differences are not significative. Regarding \cite{bunescu2006subsequence}, all the combinations are able to
significantly outperform it in terms of $F_1$.

\section{Conclusions and Future Work}
\label{conclusion}

This paper proposes a solution for Relationship Extraction (RE) based on labeled graphs kernels. The proposed kernel
is a particularization of the Random Walk Kernel for generic labeled graphs presented in \cite{Kashima03marginalizedkernels}.
In order to make the kernel suitable for RE tasks, we exploited two properties typically used in this line of
work: \textit{(i)} the words between the candidate entities or connecting them in a
syntactic representation are particularly likely to carry information regarding the relationship; and \textit{(ii)} 
combining information from distinct sources in a  kernel may help the RE system to make better decisions. Our experiments
show that the performance of our solution is comparable with the state-of-the-art on RE. Moreover, we showed
that combining our solution with other methods for RE leads to significant gains in terms of performance.

Interesting topics for future work include the study of different parameterizations of the Random Walk Kernel
for RE. Namely, we want to try different kernels for vertex and edge labels as well as different probability distributions
associated to the vertexes and the transitions. Moreover, it would be interesting to compare this kernel directly with
other methods and test the combination of other kernels with ours. Finally, we would also like to test our solution with
other datasets, namely the ACE dataset, which is composed by documents containing a wide variety of relationships (e.g., $CEO\_OF$,
$Located\_In$) involving several types of entities (e.g., $person$, $organization$, $location$).

\bibliographystyle{abbrv}
\bibliography{sigproc} 

\balancecolumns
\end{document}